\documentclass[conference, letter]{IEEEtran}
\IEEEoverridecommandlockouts
\usepackage{cite}
\usepackage{amsmath,amssymb,amsfonts}
\usepackage{algorithmic}
\usepackage{graphicx}
\usepackage{textcomp}
\usepackage{xcolor}
\def\BibTeX{{\rm B\kern-.05em{\sc i\kern-.025em b}\kern-.08em
    T\kern-.1667em\lower.7ex\hbox{E}\kern-.125emX}}

\usepackage{stmaryrd}
\renewcommand{\mod}[1]{\llbracket #1 \rrbracket}

\usepackage{balance}

\begin{document}

\title{First Order Logic with Fuzzy Semantics for Describing and Recognizing Nerves in Medical Images\\
\thanks{This work was partially supported by I. Bloch's chair in AI (Sorbonne Universit\'e and SCAI), and by projects from Institut Polytechnique
de Paris, Institut Imagine, and Ligue contre
le Cancer.}
}


\author{\IEEEauthorblockN{1\textsuperscript{st} Isabelle Bloch}
\IEEEauthorblockA{\textit{Sorbonne Université,} \\
\textit{CNRS, LIP6}\\
Paris, France \\
isabelle.bloch@sorbonne-universite.fr}
\and\IEEEauthorblockN{2\textsuperscript{nd} Enzo Bonnot}
\IEEEauthorblockA{\textit{LTCI, Télécom Paris} \\
\textit{Institut Polytechnique de Paris}\\
Paris, France \\
enzo.bonnot@telecom-paris.fr}
\and
\IEEEauthorblockN{3\textsuperscript{rd} Pietro Gori}
\IEEEauthorblockA{\textit{LTCI, Télécom Paris} \\
\textit{Institut Polytechnique de Paris}\\
Paris, France \\
pietro.gori@telecom-paris.fr}
\and
\IEEEauthorblockN{4\textsuperscript{th} Giammarco La Barbera}
\IEEEauthorblockA{\textit{IMAG2, Institut Imagine} \\
\textit{Université Paris Cité}\\
Paris, France \\
giammarco.labarbera@institutimaging.org}
\and
\IEEEauthorblockN{5\textsuperscript{th} Sabine Sarnacki}
\IEEEauthorblockA{\textit{Service de chirurgie pédiatrique} \\
\textit{Université Paris Cité, Hôpital Necker Enfants-Malades, APHP}\\
Paris, France \\
sabine.sarnacki@aphp.fr}
}

\maketitle

\begin{abstract}
This article deals with the description and recognition of fiber bundles, in particular nerves, in medical images, based on the anatomical description of the fiber trajectories. To this end, we propose a logical formalization of this anatomical knowledge. The intrinsically imprecise description of nerves, as found in anatomical textbooks, leads us to propose fuzzy semantics combined with first-order logic. We define a language representing spatial entities, relations between these entities and quantifiers. A formula in this language is then a formalization of the natural language description. The semantics are given by fuzzy representations in a concrete domain and satisfaction degrees of relations. Based on this formalization, a spatial reasoning algorithm is proposed for segmentation and recognition of nerves from anatomical and diffusion magnetic resonance images, which is illustrated on pelvic nerves in pediatric imaging, enabling surgeons to plan surgery.
\end{abstract}

\begin{IEEEkeywords}
Logic, fuzzy semantics, spatial reasoning, fiber bundles, segmentation, recognition.
\end{IEEEkeywords}

\section{Introduction}
Identifying nerve fibers prior to surgery enables surgeons to plan and control the procedure so as to minimize nerve damage, particularly in minimally invasive surgery where the visible operating field is very limited. Acquiring images of the patient prior to surgery is a common practice that facilitates this planning.
Magnetic resonance imaging (MRI) offers several acquisition modalities, providing information on anatomy and structure (typically T1 and T2 weighted images), function, angiography and diffusion (dMRI). dMRI enables fibers to be detected using tractography algorithms, exploiting the fact that fiber bundles cause anisotropic diffusion of water. These algorithms produce very large sets of fibers, up to several millions, among which it is very difficult to identify, and therefore segment and recognize, the fiber bundles of interest.

Approaches to achieving this identification have been proposed for white matter fibers in the brain, either by exploiting the large amount of data in this field, for example with statistical clustering methods~\cite{Garyfallidis, Guevara}, or with structural approaches exploiting anatomical knowledge~\cite{Delmonte, Wassermann}. Very little work has been done outside the brain, in other parts of the body. However, the medical and surgical value of nerve imaging has been widely demonstrated (see, for example,~\cite{Bertrand, Haakma, Lemos, Zhan, Zijta}), but techniques for segmenting and visualizing these nerves remain poorly reproducible, as they most often rely on manually given regions of interest to guide tractography.

In a preliminary study~\cite{Muller}, inspired by the structural approaches of~\cite{Delmonte, Wassermann}, we demonstrated that spatial relationships between nerves and anatomical structures allowed formalizing the description of nerve pathways in the pelvis, and proved useful for the segmentation and recognition of the nerves.

\begin{figure}[htbp]
\centerline{\includegraphics[width=\linewidth]{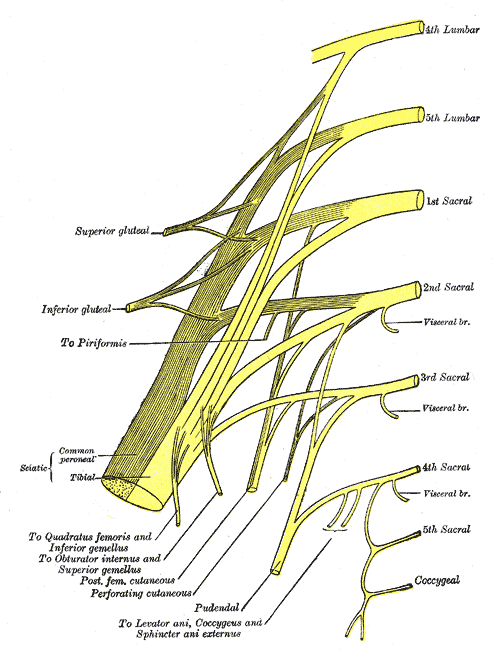}}
\caption{Sacral plexus illustration according to Gray~\cite{Gray}, plate 828.}
\label{fig:plexusSacre}
\end{figure}

Consider, for example, the sacral plexus (Fig.~\ref{fig:plexusSacre}). It is made up of several nerve roots, whose paths are described in relation to vertebral canals, sacral holes and muscles. An example of a partial description is as follows: {\em the L5 root {\bf passes} through the L5 vertebral canal, AND is {\bf anterior} to the first sacral vertebra, AND is {\bf anterior} to the piriformis muscle, AND is {\bf posterior} to the ischial spine, AND is {\bf posterior} to the iliac vessels, AND is not {\bf anterior} to the obturator muscle}.

In this article, we propose a logical formalization of this approach. The intrinsically imprecise description of nerves, as found in anatomical textbooks, leads us to propose fuzzy semantics combined with first-order logic. This is well suited to represent deterministic knowledge along with its imprecision. The idea is to define a language representing spatial entities, relations between these entities and quantifiers. A formula in this language is then a formalization of the natural language description, as given above. The semantics are given by fuzzy representations in a concrete domain~\cite{Hudelot} and satisfaction degrees of the relationships. This formalization leads to spatial reasoning algorithms for the segmentation and identification of nerves from anatomical and dMRI images, which are illustrated on pelvic nerves in pediatric imaging.

The proposed logic is a first-order fuzzy logic, such as those described in~\cite{CintulaHajek, EstevaGodoNoguera, Hajek, Novak}. Note that other logics could be used as well such as fuzzy description logic~\cite{OWL,Hudelot}. We briefly describe the syntax in Section~\ref{sec:syntax}, and propose a semantics in Section~\ref{sec:semantics}. We refer the reader to the references cited above for more a complete description of the logic, axioms, inference rules, correctness and completeness results. The main originality of our modeling lies in the semantics, and the particular modeling we propose. In Section~\ref{sec:appli}, we give some examples of application to nerve recognition in the pelvis, illustrating the usefulness of the proposed approch, on both adult and child data.

\section{Syntax}\label{sec:syntax}

The language is composed of:
\begin{itemize}
\item variables $x, y,...$ which represent regions of space (portions of fibers, anatomical regions, structuring elements used to define spatial relationships...);
\item symbols $a, b...$ for truth values, in a residuated lattice $L$, defined below;
\item functions $f, g...$, transforming a term (which may simply be a variable) into another term (e.g. the  distance function to a region produces another region);
\item the set $T$ of terms $t_i$ is the set of variables, closed by the functions; 
\item predicates $p, q...$, defining the nature of the variables (and more generally of the terms) and the relationships between the variables (spatial relationships, for example); some of these predicates allow us to take into account the polysemy of “and”, that we will detail below, by adding the sequential definition of the segments or points making up a fiber and their order along the fiber (predicate ``precedes'' or ``follows'');
\item usual logical connectives $\wedge, \vee, \rightarrow, \neg$, the conjunction $\&$ (strong conjunction) being adjoint to $\rightarrow$; 
\item $\bot, \top$;
\item quantifiers $\forall, \exists$, as well as generalized quantifiers $Q_j$ such as ``in majority'' or ``most''.
\end{itemize}

We denote by $\Phi$ the set of well-formed formulas in this language. Typically, a formula can represent a fiber, or a bundle of fibers (a nerve root). We denote by $F$ the set of functions and by $P$ the set of predicates. The set of formulas is constructed inductively from variables, the application of functions and predicates, all formulas of the type $\forall x \varphi$, $\exists x \varphi$, $Q_j \varphi$, and all those constructed with logical connectives.

Note that the negation $\neg$ is introduced in the list of connectives for convenience, but it is deduced from the implication, and for any formula $\varphi$, $\neg \varphi$ is $\varphi \rightarrow \bot$.

The example given in the introduction shows that ``and'' can have several meanings. It can mean a conjunction in the classical sense, typically of two predicates that should hold for a same object (segment of a fiber). It can also have a different meaning, in particular when pertaining implicitly to different segments of a fiber. In the example given in the introduction, the L5 root is described as anterior to the piriformis muscle and not anterior to the obturator muscle. However, these two muscles are not at the same anatomical level, the latter being lower than the former. The ``and'' then refers to the path of the nerve, which passes first in front of the first muscle, then behind the second. So we are talking about two different fiber portions, each verifying one of the two relationships. A more precise description is: one segment of the nerve is anterior to the piriformis muscle, and then another segment of the nerve is not anterior to the obturator muscle. This is why language variables represent points or fiber segments (or other anatomical structures), and specific predicates are used to indicate that one segment follows the other.

The set of truth values is chosen as a residuated lattice, as in most fuzzy logics. Here we take 
$$L = ([0,1], \leq, \max, \min, C, I, 1, 0),$$ 
where $\leq$ is the usual order, $C$ a fuzzy conjunction (here a left-continuous t-norm)\footnote{The t-norms being all smaller than the min, this justifies the name strong conjunction used for $\&$, the semantics of which is $C$, as described in Section~\ref{sec:semantics}.} and $I$ the associated residual implication:
$$\forall (\alpha, \gamma) \in [0,1]^2, I(\beta, \gamma) = \sup\{\alpha \in [0,1] \mid C(\alpha,\beta) \leq \gamma \},$$ 
i.e. $C$ and $I$ are adjoint: 
$$\forall (\alpha, \beta, \gamma) \in [0,1]^3, C(\alpha, \beta) \leq \gamma \mbox{ if and only if } \alpha \leq I(\beta, \gamma).$$

\section{Semantics}\label{sec:semantics}

We define a structure $M = (D, I_F, I_P)$, where $D$ is the domain, $I_F$ the interpretation function for functions, and $I_P$ the interpretation function for predicates.
Here, we define $D = \mathcal{F}(\mathcal{S}) \cup [0,1]$, where $\mathcal{S}$ denotes the spatial domain (in practice $\mathbb{N}^3$ for a three-dimensional digital image) and $\mathcal{F}(\mathcal{S})$ the set of fuzzy subsets of $\mathcal{S}$. Therefore, the domain includes both regions or fuzzy regions of space, and satisfaction degrees (or predicates typically) in $[0,1]$.
For a function $f \in F$, the interpretation $I_F(f)$ is a function from $\mathcal{F}(\mathcal{S})^n$ into $\mathcal{F}(\mathcal{S})$, where $n$ is the arity of $f$.
For a predicate $p \in P$, the interpretation $I_P(p)$ is a function from $\mathcal{F}(\mathcal{S})^n$ into $[0,1]$, where $n$ is the arity of $p$.

In this structure, we define an evaluation $v$ that associates any variable in the language with an element of $D$ (in this case, a fuzzy subset of the spatial domain, i.e. an element of $\mathcal{F}(\mathcal{S})$). 
For $\lambda \in \{\min, \max, I, t\}$, and $(\mu, \nu) \in \mathcal{F}(\mathcal{S})^2$, for simplicity $\lambda(\mu, \nu)$ denotes the fuzzy set associating with any point $k$ of $\mathcal{S}$ the membership degree $\lambda(\mu(k), \nu(k))$.

Classically, the evaluation of any formula is defined by induction, with:
\begin{IEEEeqnarray*}{rCl}
\mod{x}_{M,v} &=& v(x)\\
\mod{f(t_1, ... t_n)}_{M,v} &=& I_F(f)(\mod{t_1}_{M,v}, ... \mod{t_n}_{M,v})\\
\mod{p(t_1, ... t_n)}_{M,v} &=& I_P(p)(\mod{t_1}_{M,v}, ... \mod{t_n}_{M,v})\\
\mod{\forall x \varphi}_{M,v} &=& \inf \{ \mod{\varphi}_{M,v'} \mid \\ && \forall y \neq x, v'(y)=v(y) \}\\
\mod{\exists x \varphi}_{M,v} &=& \sup \{ \mod{\varphi}_{M,v'} \mid \\&& \forall y \neq x, v'(y)=v(y) \}\\
\mod{\varphi \wedge \psi}_{M,v} &=& \min(\mod{\varphi}_{M,v}, \mod{\psi}_{M,v})\\
\mod{\varphi \vee \psi}_{M,v} &=& \max(\mod{\varphi}_{M,v}, \mod{\psi}_{M,v})\\
\mod{\varphi \rightarrow \psi}_{M,v} &=& I(\mod{\varphi}_{M,v}, \mod{\psi}_{M,v})\\
\mod{\varphi \& \psi}_{M,v} &=& C(\mod{\varphi}_{M,v}, \mod{\psi}_{M,v})
\end{IEEEeqnarray*}

The semantics of generalized quantifiers are defined by fuzzy quantifiers such as ``about $n$'', ``most'', ``in the majority''. For example, the semantics of ``about $n$'' is classically given by a triangular fuzzy set of modal value $n$.

\medskip

We now give some examples of functions and predicates with their semantics, which is one of the original features of this article.

Unary predicates are used to represent the nature of variables and terms. For instance, $L5(x)$ means that segment $x$ belongs to nerve root L5.

A binary predicate $p_{follow}$ can be used to describe the order of fiber segments: $p_{follow}(x,y)$ means that $x$ is located after $y$ along the fiber path (the orientation is fixed, e.g. top-down for pelvic nerves, and the reference frame, linked to the patient, is known and defined by the image acquisition process). This predicate can be seen as one of Allen's relations, drawing a parallel between the succession of fiber segments and the succession of time intervals. Its association with spatial relations is then close to the qualitative trajectory calculus proposed in~\cite{QTC}.

Morphological operators, in particular dilations, are useful to model spatial relations, as shown in~\cite{Bloch2005,BlochRalescu}. The approach described below allow modeling many types of relations, either topological ones (such as fuzzy extensions~\cite{Bloch2021} of RCC8 relations~\cite{RCC}), or metric ones (distances, directions...).
A dilation function $f_{dil}$ is then introduced to define a region of space dilated according to a given structuring element. If $t$ is a term (a variable or the result of a function) representing a region of space, and $\mu = \mod{t}_{M,v}$ its semantics, $\mu$ is an interpretation of $t$ in the concrete domain $D$, i.e. a fuzzy set defined on $\mathcal{S}$. Similarly, a structuring element is a term $t'$ whose semantics is $\nu = \mod{t'}_{M,v}$. The semantics of the dilation function is then $\mod{f_{dil}(t,t')}_{M,v} = \delta(\mu,\nu)$ where $\delta$ is the fuzzy dilation, defined as~\cite{Bloch2009}:
$$\forall k \in \mathcal{S}, \delta(\mu,\nu)(k) = \sup \{ C(\nu(k-k'), \mu(k')) \mid k'\in \mathcal{S} \}.$$
This function is very useful for defining predicates for spatial relationships between regions.

Directional relations such as ``anterior'' are widely used in the anatomical description of nerves. The proposed logic allows modeling them using functions and predicates, with interpretations based on the fuzzy modeling of spatial relations~\cite{Bloch2005,BlochRalescu}. We define a structuring element $t'$ associated with the considered direction, whose semantics $\nu$ is a fuzzy set in the spatial domain. The specific form of this structuring element depends on parameters, in particular the direction and the level of imprecision attached to the relation. In Section~\ref{sec:appli}, the directional structuring elements are defined as fuzzy cones, in the desired direction and whose aperture can be adjusted by the user if required. The function $f_{dil}(t,t')$ is a term representing the region located in the considered direction of $t$, and its semantics is a fuzzy region of space, where the value at each point represents the degree to which that point is in the direction $\nu= \mod{t'}_{M,v}$ of $\mu = \mod{t}_{M,v}$. Let $p_{dir}$ be the predicate representing the directional relationship between two variables or terms. We then define the degree to which $t''$ is in the considered direction with respect to $t$ by 
$$\mod{p_{dir}(t,t',t'')}_{M,v} = ag\{ C(\delta(\mu,\nu)(k), \xi(k)) \mid k \in \mathcal{S}\},$$ 
where $ag$ is an aggregation function with co-domain $[0,1]$, and $\xi = \mod{t''}_{M,v}$.

\begin{figure}[htbp]
\centerline{\includegraphics[width=\linewidth]{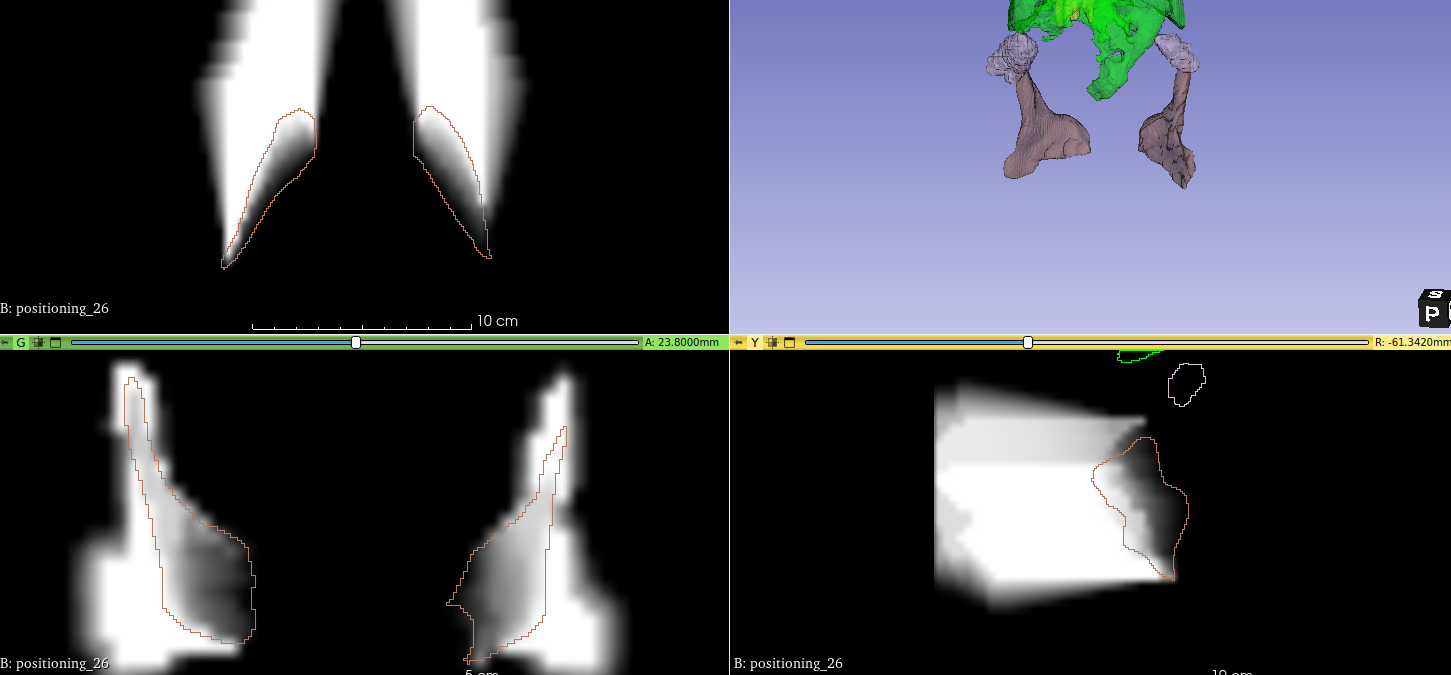}}
\caption{Interpretation in the concrete domain of the region anterior to the obturator muscle (muscle contours are shown in red). Slices in three orthogonal directions are displayed.}
\label{fig:anteriorObturator}
\end{figure}

Fig.~\ref{fig:anteriorObturator} illustrates the interpretation, in the concrete domain defined by the image, of the result of the function $f_{anterior}$ applied to the obturator muscle. The muscle has been removed from the dilation, except near its contour, to account for spatial imprecision in muscle segmentation and in the registration between anatomical and diffusion images. 

The L5 root is described by a succession of segments represented by variables $x_1, ... x_n...$, where $x_n$ is the segment located behind the obturator muscle (i.e. not anterior to it). This segment must satisfy the formula 
$$L5(x_n) \equiv (\forall x_i, i=1...n-1, L5(x_i) \& p_{follow}(x_{i+1}, x_i)) $$
$$\& \neg p_{dir}(OM, ANT, x_n),$$
where $OM$ is the term corresponding to the obturator muscle, and $ANT$ the one corresponding to the relation. The semantics of this formula allows us to manipulate fuzzy regions of space, representing, respectively, the muscle (segmented from an anatomical MRI image), the semantics of ``anterior'', defining a fuzzy structuring element, the region anterior to the muscle (calculated by dilating the muscle with this structuring element), and the fiber segments. Their combination provides a satisfaction degree of the set of constraints for a fiber segment.

\begin{figure}[htbp]
\centerline{\includegraphics[width=\linewidth]{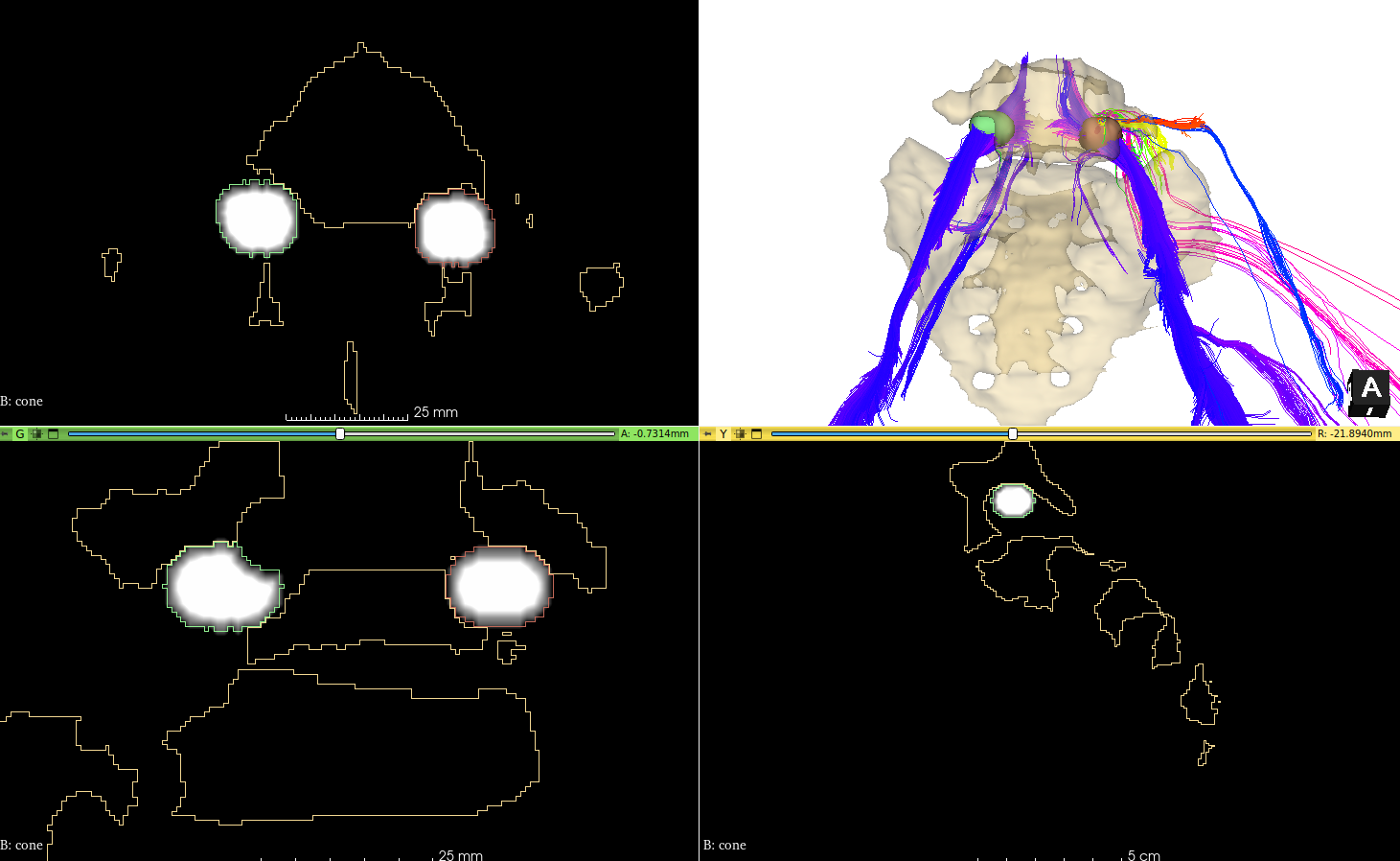}}
\caption{Interpretation of the predicate ``crossing'' in the concrete domain.}
\label{fig:crossing}
\end{figure}

In a similar way, we define predicates representing distance relations, ternary relations such as ``between'' and ``in the middle''. The semantics of the predicate ``crossing'' or ``goes through'' ({\em L5 root {\bf goes through} vertebral canal L5}) is defined from a distance, with a value increasing from 0 at the edge of the topological loop of the object (the vertebral canal in the example) to a maximal value at its center. This is illustrated in Fig.~\ref{fig:crossing}. 

When a fiber bundle starts or ends in a particular structure, predicates representing this connectedness are modeled using generalized quantifiers such as ``about $n$'', whose semantics are given by fuzzy numbers.

\section{Application to pelvic nerves recognition}\label{sec:appli}

\begin{figure}[htbp]
\centerline{\includegraphics[width=0.5\linewidth]{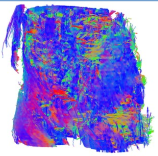}}
\caption{Example of the whole set of fibers computed from dMRI using a tractography algorithm.}
\label{fig:chignon}
\end{figure}

In this section, we illustrate how the proposed logic is applied to the segmentation and recognition of pelvic nerves. For each patient, anatomical structures are segmented in anatomical T2 MRI (the description of the method is beyond the scope of this paper). All fibers are obtained from diffusion MRI using a deterministic tractography algorithm~\cite{Tournier}. This provides a huge set of fibers, as in Fig.~\ref{fig:chignon}, from which we want to extract the meaningful ones, corresponding to specific nerves of interest.

Anatomical definitions of nerves and their pathways are formalized in the proposed logic. Based on these descriptions and their fuzzy semantics, the segmentation and recognition algorithm is a spatial reasoning algorithm that relies on the translation of this knowledge in the form of queries, consisting mainly of a succession of spatial relations to be checked. The degrees of satisfaction of each predicate are computed and aggregated, then the result is thresholded to make the final decision. Alternatively, each predicate value could be thresholded individually before being aggregated.

Using this approach, we were able to describe a fiber belonging to the L5 nerve bundle as any fiber satisfying the following query (the values of the relationship parameters and thresholds are not mentioned here for the sake of readability):

\begin{tabular}{lp{6cm}}
L5 $=$ & crossing(VertebralCanalL5) \\
& then  anterior\_of(PiriformisMuscles) \\
& then (inferior\_of(PiriformisMuscles) and \\ & \qquad lateral\_of(LevatorAniMuscles))\\
& then not posterior\_of(VertebraL5)\\
& then not posterior\_of(Sacrum)\\
& then not crossing(SacralHoleS1) \\
& then not posterior\_of(PiriformisMuscles) \\
& then not anterior\_of(ObturatorMuscles)\\
& then not between(LeftObturatorMuscle, \\ & \qquad Right\-ObturatorMuscle)
\end{tabular}

A fiber belonging to S1 satisfies the following query:

\begin{tabular}{lp{6cm}}
S1 = & crossing(SacralHoleS1) \\
&then anterior\_of(PiriformisMuscles)\\
&then (inferior\_of(PiriformisMuscles) and \\ & \qquad lateral\_of(LevatorAniMuscles))\\
&then not posterior\_of(VertebraL5)\\
&then not posterior\_of(Sacrum)\\
&then not (crossing(VertebralCanalL5) or  \\ & \qquad crossing(SacralHoleS2))\\
&then not posterior\_of(PiriformisMuscles)\\
&then not anterior\_of(ObturatorMuscles)\\
&then not between(LeftObturatorMuscle, \\ & \qquad Right\-ObturatorMuscle)
\end{tabular}

The following query isolates the S2 root fibers:

\begin{tabular}{lp{6cm}}
S2 = & crossing(SacralHoleS2)\\
&then anterior\_of(PiriformisMuscles)\\
&then (inferior\_of(PiriformisMuscles) and \\ & \qquad lateral\_of(LevatorAniMuscles))\\
&then not posterior\_of(VertebraL5)\\
&then not posterior\_of(Sacrum)\\
&then not (crossing(SacralHoleS1) or  \\ & \qquad crossing(SacralHoleS3))\\
&then not lateral\_of(PiriformisMuscles)\\
&then not anterior\_of(ObturatorMuscles)\\
&then not between(LeftObturatorMuscle, \\ & \qquad Right\-ObturatorMuscle)
\end{tabular}

Finally, the query for recognizing the S3 nerve root is written as:

\begin{tabular}{lp{6cm}}
S3 = & crossing(SacralHoleS3)\\
&then anterior\_of(PiriformisMuscles)\\
&then (inferior\_of(PiriformisMuscles) and \\ & \qquad lateral\_of(LevatorAniMuscles))\\
&then not posterior\_of(VertebraL5)\\
&then not posterior\_of(Sacrum)\\
&then not (crossing(SacralHoleS2) or \\ & \qquad crossing(SacralHoleS4))\\
&then not lateral\_of(PiriformisMuscles)\\
&then not anterior\_of(ObturatorMuscles)\\
&then not between(LeftObturatorMuscle, \\ & \qquad Right\-ObturatorMuscle)
\end{tabular}

\medskip

In each of the queries described, the term ``then'' corresponds to the ``and'' referring to the nerve path satisfying first one relationship and then the next, which we have modeled using the predicate $p_{follow}$. This introduces an order of validation of spatial relations for considering a fiber as belonging to the bundle we are looking for. 

\begin{figure}[htbp]
\centerline{\includegraphics[width=\linewidth]{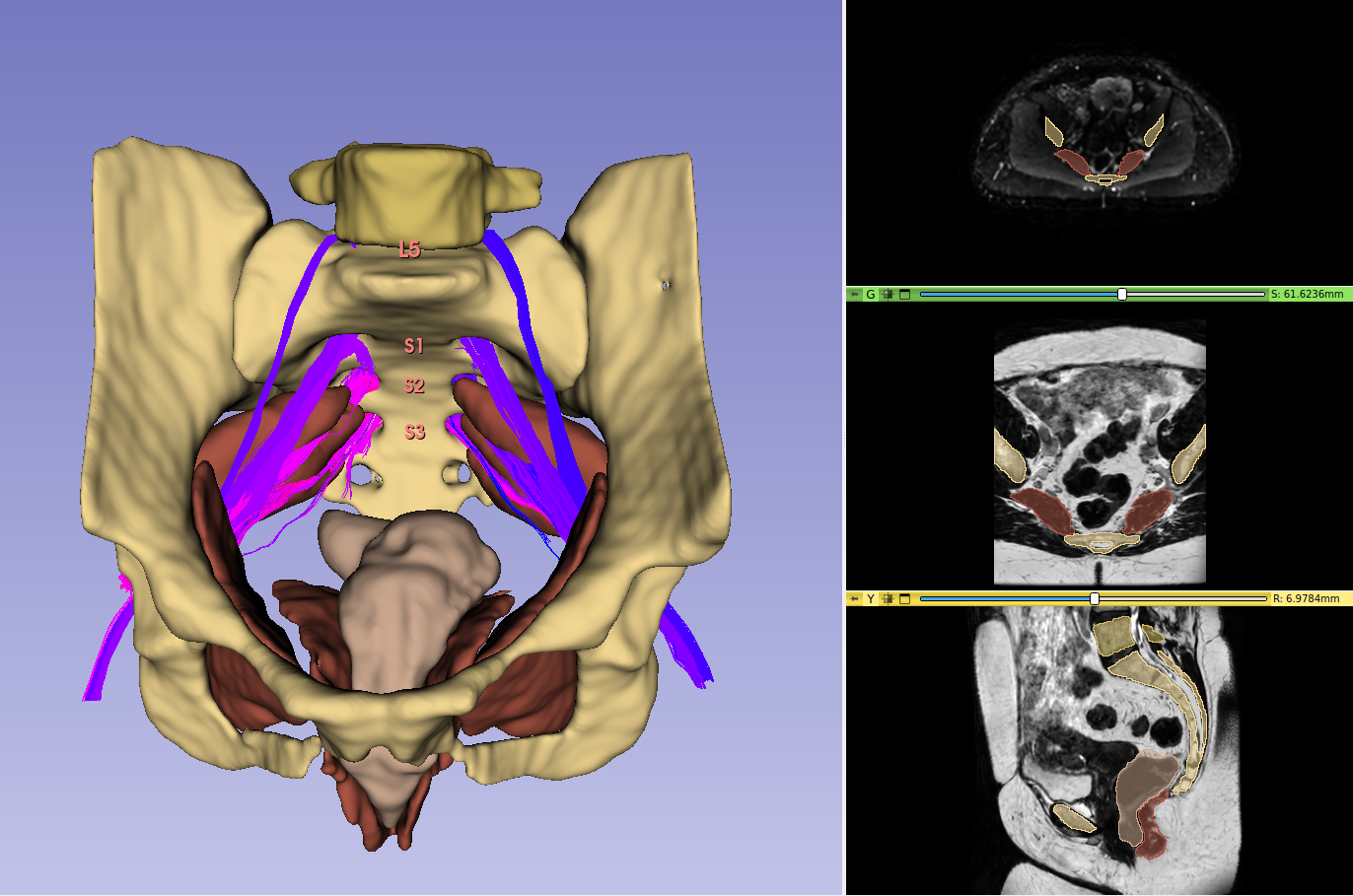}}
\caption{Illustration of the L5-S3 pelvic fiber recognition result for a healthy adult. Left: 3D representation. Right, from top to bottom: dMRI axial slice, T2 MRI axial slice, T2 MRI sagittal slice.}
\label{fig:plexusSacre3D}
\end{figure}

\begin{figure}[htbp]
\centerline{\includegraphics[width=\linewidth]{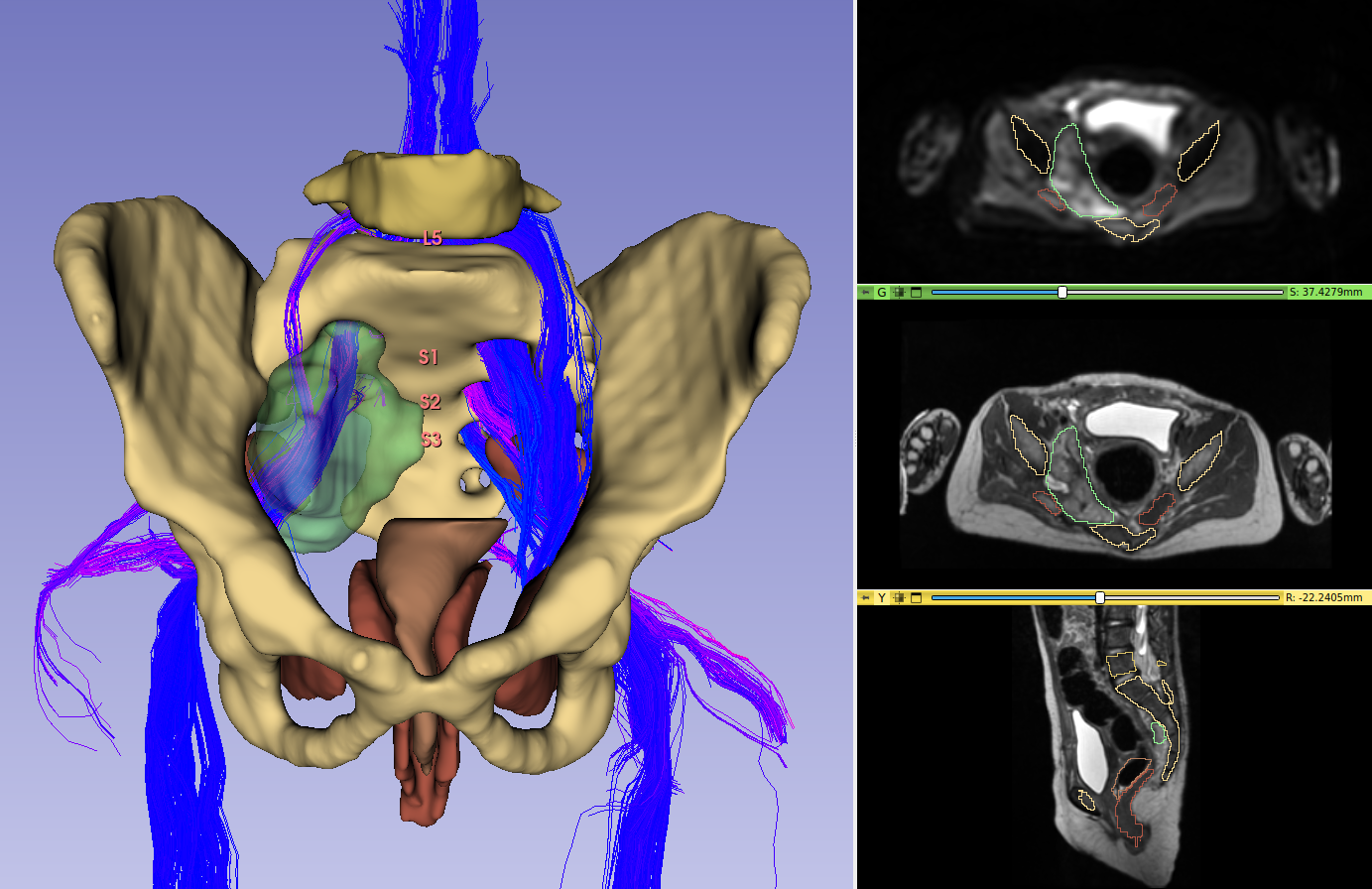}}\caption{Illustration of the L5-S3 pelvic fiber recognition result for a child with a tumor (in green). Left: 3D representation.  Right, from top to bottom: dMRI axial slice, T2 MRI axial slice, T2 MRI sagittal slice.}
\label{fig:plexusSacre3D2}
\end{figure}

Fig.~\ref{fig:plexusSacre3D} shows a result in a healthy adult subject, and Fig.~\ref{fig:plexusSacre3D2} in a pathological child subject. In both cases, the nerves are well segmented and well recognized, even in the pediatric case, which is much more difficult due to the small size of the structures. These good results, even in pathological cases, are the result of the proposed fuzzy modeling, allowing anatomical variability to be taken into account, and of the combination of several relationships before making a decision. This guarantees a good robustness to pathologies. Last but not least, the 3D visualizations provided are a useful aid to the surgeon~\cite{Muller}.

\section{Conclusion}

In this paper we have proposed a fuzzy logic formalization of anatomical knowledge about nerves. In particular, we have proposed interpretations in the concrete image domain of functions and degrees of satisfaction of predicates, which make it possible to represent the spatial relationships between fiber segments and anatomical structures, as well as the succession of these segments along the fiber path. This formalization makes it possible to reason in image space, and has given rise to an algorithm for query-based nerve segmentation and recognition. Illustrations of pelvic nerves demonstrate the value of this approach for visualizing nerves in a 3D digital twin of the patient, useful for surgical planning.

The proposed method exploits available knowledge, links it to data without the need for heavy learning procedures, and is directly explainable, via the pieces of knowledge actually involved in the final decision.

Our ongoing and future work is twofold: on the one hand, to continue formalizing and detailing the semantics, with the associated calculations and algorithms, and on the other hand, to extend the application fields to other nerve roots.

%

\balance

\end{document}